\documentclass[conference]{IEEEtran}
\usepackage{cite}
\usepackage{graphicx}
\usepackage{amsmath}
\usepackage{amssymb} % For \mathbb command
\usepackage{hyperref}
\usepackage{etoolbox}
\usepackage{multirow}
\usepackage{booktabs} % For better looking tables
\usepackage{amsfonts} % For math symbols like \mathbb{R}
\usepackage{algorithm}
\usepackage{algpseudocode}

\hypersetup{colorlinks=true,linkcolor=blue,citecolor=blue,urlcolor=blue}

\begin{document}
\title{TS2Vec-Ensemble: An Enhanced Self-Supervised Framework for Time Series Forecasting}

\author{\IEEEauthorblockN{Ganeshan Niroshan}
\IEEEauthorblockA{\textit{Department of Computer Science and Engineering} \\
\textit{University of Moratuwa}\\
Colombo, Sri Lanka \\
ganeshan.21@cse.mrt.ac.lk}
\and
\IEEEauthorblockN{Uthayasanker Thayasivam}
\IEEEauthorblockA{\textit{Department of Computer Science and Engineering} \\
\textit{University of Moratuwa}\\
Colombo, Sri Lanka \\
rtuthaya@cse.mrt.ac.lk}
}

\maketitle

\begin{abstract}
Self-supervised representation learning, particularly through contrastive methods like TS2Vec, has advanced the analysis of time series data. However, these models often falter in forecasting tasks because their objective functions prioritize instance discrimination over capturing the deterministic patterns, such as seasonality and trend, that are critical for accurate prediction. This paper introduces \textbf{TS2Vec-Ensemble}, a novel hybrid framework designed to bridge this gap. Our approach enhances the powerful, implicitly learned dynamics from a pre-trained TS2Vec encoder by fusing them with explicit, engineered time features that encode periodic cycles. This fusion is achieved through a dual-model ensemble architecture, where two distinct regression heads—one focused on learned dynamics and the other on seasonal patterns—are combined using an adaptive weighting scheme. The ensemble weights are optimized independently for each forecast horizon, allowing the model to dynamically prioritize short-term dynamics or long-term seasonality as needed. We conduct extensive experiments on the ETT benchmark datasets for both univariate and multivariate forecasting. The results demonstrate that TS2Vec-Ensemble consistently and significantly outperforms the standard TS2Vec baseline and other state-of-the-art models, validating our hypothesis that a hybrid of learned representations and explicit temporal priors is a superior strategy for long-horizon time series forecasting.
\end{abstract}


\section{Introduction}
\label{sec:introduction}

\subsection{Background and Motivation}
Time series forecasting remains a critical task across numerous scientific and industrial domains, including financial market prediction, energy consumption management, and climate modeling. A widely adopted benchmark for evaluating forecasting models is the Electricity Transformer Temperature (ETT) dataset, which comprises sensor readings from power transformers and presents complex, real-world temporal dynamics \cite{zhou2021informer}. While traditional statistical methods such as ARIMA have long served as foundational baselines, their inherent assumptions of linearity and stationarity restrict their efficacy on data characterized by intricate nonlinear dependencies \cite{box2015time}. Consequently, the field of Long-Term Time Series Forecasting (LTSF) has increasingly shifted towards deep learning architectures like Recurrent Neural Networks (RNNs) and Transformers, which possess the capacity to autonomously learn long-range, complex patterns from raw data.

A significant recent advancement in this area is the application of self-supervised learning (SSL) to learn powerful representations from unlabeled time series. Among these, **TS2Vec** has emerged as a state-of-the-art framework, employing a hierarchical contrastive learning objective to produce representations that preserve both local and global temporal consistency \cite{yue2022ts2vec}. The success of TS2Vec demonstrates a strong capacity for generalization and robustness across diverse time series tasks.

Despite its strengths in representation learning, the direct application of TS2Vec to forecasting reveals a critical limitation: its purely contrastive objective does not explicitly model deterministic components such as seasonality and trend. These components are often the primary drivers of predictability in long-horizon forecasting. This research aims to bridge this capability gap by proposing a novel framework that strategically integrates explicit temporal features to enhance TS2Vec's forecasting performance.

\subsection{Research Contributions}
To address the aforementioned limitations, this study proposes an enhanced forecasting framework built upon the TS2Vec architecture. Our primary contributions are threefold:
\begin{enumerate}
    \item \textbf{Ablation of Hybrid Objectives:} We conduct a systematic investigation into augmenting TS2Vec with a masked signal modeling (MSM) objective and a separate hybrid decomposition model using XGBoost. Our analysis exposes the fundamental limitations of these approaches, thereby motivating the development of a more sophisticated integration strategy.
    
    \item \textbf{An Enhanced TS2Vec Ensemble Framework:} We introduce a novel ensemble methodology that fuses the rich, implicit embeddings from TS2Vec with explicit, engineered sinusoidal time features. This framework utilizes two distinct regression heads and an adaptive, validation-optimized weighting mechanism to dynamically balance the influence of learned dynamics against deterministic seasonal patterns for each forecast horizon.
    
    \item \textbf{Comprehensive Empirical Validation:} We perform extensive experiments on the ETT benchmark datasets for both univariate and multivariate forecasting tasks. Our empirical results demonstrate that the proposed TS2Vec-Ensemble consistently and significantly outperforms the original TS2Vec baseline and other state-of-the-art models, particularly in challenging long-horizon scenarios.
\end{enumerate}

\section{Related Work}

\subsection{Classical and Modern Forecasting Models}
Time series forecasting has a rich history rooted in statistical methods like the Autoregressive Integrated Moving Average (ARIMA) model, which excels at modeling linear relationships in stationary time series by differencing away trends and seasonality \cite{box2015time}. While foundational, these methods are often insufficient for the complex, nonlinear patterns found in modern datasets. This led to the adoption of deep learning, initially with Recurrent Neural Networks (RNNs) and their variants like LSTMs, which naturally model sequential dependencies.

More recently, architectures inspired by advances in other domains have pushed the boundaries of Long-Term Time Series Forecasting (LTSF). Transformer-based models, such as **Informer** \cite{zhou2021informer}, Autoformer, and FEDformer, adapted the self-attention mechanism to capture long-range dependencies more efficiently than RNNs. However, the surprising effectiveness of simpler models like **DLinear**—which uses a basic decomposition and linear layers—has challenged the notion that complexity is always necessary \cite{zeng2023are}. Other innovative approaches include **PatchTST**, which improves performance by treating subseries-level "patches" as input tokens \cite{nie2023time}, and **TimesNet**, which captures multi-periodicity by converting the 1D series into a 2D tensor for analysis with standard computer vision backbones.

\subsection{Unsupervised and Self-Supervised Representation Learning}
Parallel to supervised methods, learning effective representations from unlabeled time series has become a major focus. Early deep learning approaches included autoencoders, like the one used in TimeNet, which learn representations by reconstructing the input signal. Other methods focused on preserving specific properties, such as pairwise similarities in the time domain, or using randomized kernels to generate embeddings.

The paradigm of contrastive learning, popularized in computer vision, has been particularly influential. This approach trains a model to pull representations of "positive pairs" (augmented views of the same instance) closer together while pushing apart "negative pairs" (representations from different instances). Several methods have adapted this to time series:
\begin{itemize}
    \item \textbf{T-Loss} learns scalable representations for multivariate time series using a triplet loss with time-based negative sampling. It primarily focuses on instance-level representations \cite{yue2022ts2vec}.
    \item \textbf{TS-TCC} encourages consistency between strongly and weakly augmented versions of a time series to learn transformation-invariant features \cite{yue2022ts2vec}.
    \item \textbf{TNC (Temporal Neighborhood Coding)} leverages the local smoothness of a signal, defining temporal neighborhoods to learn representations that are robust to non-stationarity \cite{yue2022ts2vec}.
\end{itemize}
While powerful, these methods often learn representations at a single, fixed semantic level and rely on strong assumptions (e.g., transformation invariance) that may not hold for all time series data \cite{yue2022ts2vec}.

\subsection{TS2Vec: Hierarchical Contrastive Learning}
Our work builds directly upon **TS2Vec**, a universal framework that addresses many of the limitations of prior methods \cite{yue2022ts2vec}. Unlike models that generate a single instance-level representation, TS2Vec learns fine-grained, timestamp-level representations. Its core innovation is **hierarchical contrastive learning**, which captures multi-scale contextual information by applying contrastive losses at both the temporal and instance levels across different resolutions \cite{yue2022ts2vec}.

Instead of relying on transformations, TS2Vec introduces **contextual consistency**. It generates positive pairs by taking the representation of the same timestamp from two different "context views" of the series, which are created via random cropping and a novel timestamp masking technique applied to a latent projection of the input \cite{yue2022ts2vec}. This allows the model to learn robust contextual representations without imposing potentially unsuitable inductive biases like transformation invariance. The resulting representations have proven highly effective, achieving state-of-the-art results not only in classification but also in downstream tasks like forecasting and anomaly detection with simple linear models trained on top \cite{yue2022ts2vec}. Our work aims to further enhance the forecasting capabilities of this powerful foundation.

\section{Foundational Architecture: TS2Vec}

\subsection{Encoder Architecture}
Our enhancements are built upon the robust TS2Vec encoder, which is designed to generate fine-grained representations for each timestamp. The encoder consists of three main components:
\begin{enumerate}
    \item \textbf{Input Projection Layer:} A fully connected layer maps the raw observation at each timestamp to a high-dimensional latent vector. This is a critical step as time series values can be unbounded, making it impossible to define a universal mask token for raw data.
    \item \textbf{Timestamp Masking:} To generate augmented context views, the model randomly masks latent vectors at selected timestamps.
    \item \textbf{Dilated CNN Module:} A powerful backbone of ten residual blocks, with a hidden dimension of 64 and producing a final representation dimension of 320, extracts the contextual representation. Each block contains 1-D dilated convolutions, with the dilation parameter increasing exponentially in deeper blocks to create a large receptive field.
\end{enumerate}

\begin{figure*}[t]
\centering
\includegraphics[width=0.8\textwidth]{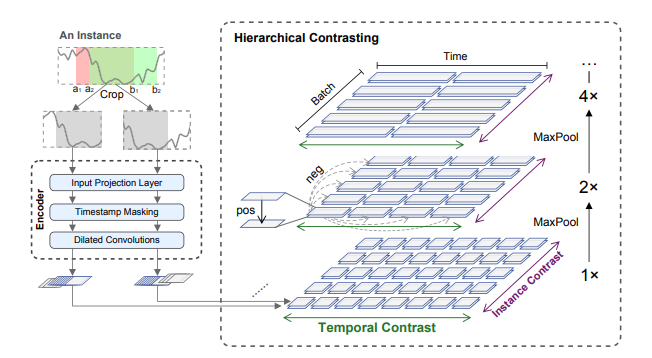}
\caption{The original TS2Vec architecture with hierarchical contrastive learning. The encoder uses multi-scale dilated convolutions and contrastive objectives to learn robust representations \cite{yue2022ts2vec}.}
\label{fig:ts2vec_architecture}
\end{figure*}

% Double column equation
\begin{figure*}[!t]
\normalsize
\begin{equation} \label{eq:temp_loss}
\mathcal{L}_{\text{temp}}^{(i,t)} = -\log\frac{\exp(r_{i,t} \cdot r'_{i,t})}{\sum_{t' \in \Omega} \left( \exp(r_{i,t} \cdot r'_{i,t'}) + \mathbf{1}_{[t \neq t']} \exp(r_{i,t} \cdot r_{i,t'}) \right)}
\end{equation}
\hrulefill
\end{figure*}

% Double column equation
\begin{figure*}[!t]
\normalsize
\begin{equation} \label{eq:inst_loss}
\mathcal{L}_{\text{inst}}^{(i,t)} = -\log\frac{\exp(r_{i,t} \cdot r'_{i,t})}{\sum_{j=1}^{B} \left( \exp(r_{i,t} \cdot r'_{j,t}) + \mathbf{1}_{[i \neq j]} \exp(r_{i,t} \cdot r_{j,t}) \right)}
\end{equation}
\hrulefill
\end{figure*}

\subsection{Contextual Consistency and Hierarchical Contrasting}
The training objective of TS2Vec is centered around two novel concepts:
\textbf{Contextual Consistency:} This strategy posits that the representations of the same timestamp should be consistent across two different, augmented views of the input series, which are generated via timestamp masking and random cropping.
\textbf{Hierarchical Contrasting:} To capture multi-scale information, TS2Vec performs contrastive learning at multiple scales. A dual contrastive loss—combining temporal and instance-wise losses—is applied at each level of the hierarchy, allowing the model to learn both instance-specific characteristics and dynamic trends.

\subsection{Dual Contrastive Loss Function}
The core of TS2Vec's learning process is a dual objective that leverages both temporal and instance-wise contrastive losses. This design allows the model to capture complementary aspects of the data. For example, in a dataset of electricity consumption from multiple users, the instance-wise contrast helps learn user-specific patterns, while the temporal contrast focuses on mining the dynamic trends over time.

\subsubsection{Temporal Contrastive Loss}
This loss encourages discriminative representations over time. It treats the representations of the same timestamp $t$ from two augmented views of a time series $x_i$ (denoted as $r_{i,t}$ and $r'_{i,t}$) as a positive pair, while representations at different timestamps ($t' \neq t$) are treated as negative pairs. The loss is formulated as shown in Equation~\ref{eq:temp_loss}, where $\Omega$ is the set of timestamps in the overlapping segment of the two augmented views.

\subsubsection{Instance-wise Contrastive Loss}
This loss distinguishes a specific time series instance from others in the batch. For a given timestamp $t$, the representations from the two views of the same instance $i$ ($r_{i,t}$ and $r'_{i,t}$) form a positive pair, while representations from all other instances $j \neq i$ in the batch serve as negative samples. The loss is defined as shown in Equation~\ref{eq:inst_loss}, where $B$ is the batch size.

\subsubsection{Overall Loss}
The final objective combines both losses, averaged over all instances and timestamps, and is applied at each level of the hierarchical structure:
\begin{equation}
\mathcal{L}_{\text{dual}} = \frac{1}{NT}\sum_{i}\sum_{t}(\mathcal{L}_{\text{temp}}^{(i,t)} + \mathcal{L}_{\text{inst}}^{(i,t)})
\end{equation}

\section{Proposed Methods and Architectural Ablation}

While TS2Vec's contextual consistency provides a robust, data-driven representation, it does not explicitly encode deterministic periodicities. For forecasting tasks dominated by such cycles, this can be a limitation. We therefore investigated three distinct enhancement strategies.

\subsection{Investigated Method 1: TS2Vec with Masked Signal Modeling}
\subsubsection{Dual Objective Formulation}
The TS2Vec+MSM method extends the encoder architecture by adding a shallow, lightweight decoder network (e.g., a 3-layer MLP) dedicated to reconstruction. This objective aims to enhance the encoder’s ability to capture local continuity. The model is optimized using a combined loss function that balances the contrastive learning objective ($\mathcal{L}_{\text{contrastive}}$) with the Masked Signal Modeling reconstruction objective ($\mathcal{L}_{\text{MSM}}$):
\begin{equation}
\mathcal{L} = (1-\lambda)\mathcal{L}_{\text{contrastive}} + \lambda\mathcal{L}_{\text{MSM}}
\end{equation}
The reconstruction loss ($\mathcal{L}_{\text{MSM}}$) is implemented as Mean Squared Error (MSE) and is calculated exclusively on the randomly masked positions of the input signal. A dynamic warm-up schedule is employed for the weighting factor $\lambda$, which starts near zero and gradually increases. This strategy is designed to allow the model to first acquire global, discriminative representations before introducing the pressure for fine-grained, local fidelity via reconstruction.

\subsubsection{Analysis of Failure Mode}
Despite the theoretical promise, the TS2Vec+MSM approach yielded the poorest empirical results, achieving an average MSE of 0.242 and MAE of 0.404 (see Table~\ref{tab:ablation_results}). Compared to the baseline TS2Vec average (MSE 0.116 / MAE 0.253), the MSM integration resulted in nearly a doubling of the error. This substantial underperformance confirms that integrating a simplistic local reconstruction objective significantly disrupts the core functionality of the contrastive encoder. While the contrastive loss encourages scale-invariance, the reconstruction loss forces the encoder to capture high-fidelity, localized information. This forced specialization compromises the global, hierarchical structure of the TS2Vec representation, yielding embeddings that are poorly generalized for the multi-horizon regression task \cite{chen2020simple, yue2022ts2vec}. Our ablation study further supported this, showing that disabling the MSM loss ($\lambda = 0$) substantially improved this method's performance.

\subsection{Investigated Method 2: Hybrid Sinusoidal–XGBoost Decomposition}
The Boosted Hybrid model is a specialized pipeline designed to explicitly leverage classical time series features with non-linear boosting.
\subsubsection{Stage 1: Sinusoidal Regression for Periodicity}
In the initial stage, a feature set is engineered to capture deterministic time dependencies. These features include lag values, rolling statistics (mean and standard deviation), and critical Fourier features—sinusoidal and cosine components—to encode daily and weekly periodicities \cite{lai2018modeling}. A simple linear regression model is trained on these features to produce a baseline forecast, capturing the primary linear trends and seasonal components.

\subsubsection{Stage 2: XGBoost as a Residual Corrector}
The strength of this hybrid method lies in its second stage, which involves modeling the non-linear residual component (the error left by the linear model). XGBoost regressors are trained separately for each forecast horizon to predict this residual \cite{oreshkin2019n}. Gradient boosting excels at learning complex, localized interactions and non-linear remainder trends that defy the structure of the initial linear model \cite{chen2016xgboost}. Although the overall average performance of this hybrid (Avg MSE 0.245 / MAE 0.383) is low due to its reliance on manually engineered features, the model demonstrates specialized utility. For instance, in the ETTh2 H=48 forecast, it achieved an excellent MAE of 0.269 and a near-best MSE of 0.122 (Table~\ref{tab:ablation_results}). This niche performance validates the decomposition principle: a simple feature set, when combined with a powerful non-linear corrector, can accurately model specific short-term, irregular errors.

\subsection{Proposed Method: TS2Vec-Ensemble with Time Features}
Our final and most effective approach is an adaptive ensemble that fuses the complementary strengths of implicit and explicit feature modeling. Two parallel Ridge regression heads are trained:
\begin{itemize}
    \item \textbf{Model A (Dynamics-focused):} Uses only the original TS2Vec embeddings, capturing complex, data-driven dynamic patterns.
    \item \textbf{Model B (Seasonality-focused):} Uses TS2Vec embeddings concatenated with explicit sinusoidal time features. To capture daily cycles, two features are generated for each timestamp $t$: $\sin(2\pi t / 24)$ and $\cos(2\pi t / 24)$. This anchors predictions to known deterministic patterns.
\end{itemize}
The final forecast is a validation-optimized, horizon-adaptive weighted average of the two models' predictions. The weights $(\omega_1, \omega_2)$ are selected from a grid of 17 candidates (e.g., [0.9, 0.1], [0.85, 0.15], ..., [0.1, 0.9]) using a holdout validation set to prevent overfitting and find the optimal balance for each specific forecasting task.

\subsection{Algorithmic Formulation of the Ensemble Method}
The complete end-to-end process for our proposed ensemble strategy is formalized in Algorithm~\ref{alg:ensemble_method}. The algorithm is modular, consisting of representation learning, dual feature engineering, parallel model training, and a validation-based optimization of the ensemble weights for each forecast horizon. To predict a target at time $t$, we use the representation of the last available observation, $z_{t-h}$. The feature vector $\mathbf{F}_t$ is constructed from this embedding. For the enhanced model, this is concatenated with time features $\mathbf{T}_t$. The ensemble weights $\mathbf{w}^*$ are optimized by minimizing $\sqrt{MSE} + MAE$ on a validation set. This approach leverages representation complementarity: TS2Vec embeddings capture complex non-linear dynamics, while explicit time features provide a robust anchor for periodic patterns, reducing overall forecast variance.

\begin{algorithm}[t]
\caption{TS2Vec-Ensemble Forecasting}
\label{alg:ensemble_method}
\begin{algorithmic}[1]
\Require 
Time series $\mathbf{X} \in \mathbb{R}^{N \times T \times D}$; Horizons $H = \{h_1, \ldots, h_k\}$; Candidate weights $W_{cand}$
\Ensure 
Forecasts $\hat{\mathbf{Y}} = \{\hat{\mathbf{Y}}_h \mid h \in H\}$

\Statex \textit{// Phase 1: Representation Learning}
\State $\theta^* \gets \text{TrainTS2Vec}(\mathbf{X}_{\text{train}})$
\State $\mathbf{Z} \in \mathbb{R}^{N \times T \times d} \gets \text{TS2VecEncode}(\mathbf{X}, \theta^*)$

\For{each horizon $h \in H$}
    \Statex \textit{// Phase 2: Feature Engineering for each time t}
    \State $\mathbf{F}_{t, \text{orig}} \gets z_{t-h}$ \Comment{Use last timestamp embedding}
    \State $\mathbf{T}_t \gets [\sin(2\pi t/24), \cos(2\pi t/24)]^T$
    \State $\mathbf{F}_{t, \text{enh}} \gets [\mathbf{F}_{t, \text{orig}} \| \mathbf{T}_t]$ \Comment{Concatenate}
    
    \Statex \textit{// Phase 3: Train Dual Regression Models}
    \State $M_{\text{orig}} \gets \text{Ridge}(\{\mathbf{F}_{t, \text{orig}}\}_{\text{train}}, \mathbf{Y}_{\text{train}})$
    \State $M_{\text{enh}} \gets \text{Ridge}(\{\mathbf{F}_{t, \text{enh}}\}_{\text{train}}, \mathbf{Y}_{\text{train}})$
    
    \Statex \textit{// Phase 4: Optimize Ensemble Weights}
    \State $\mathbf{w}^* \gets \arg\min_{\mathbf{w} \in W_{cand}} \left( \sqrt{\text{MSE}(\hat{\mathbf{Y}}_{\text{val}})} + \text{MAE}(\hat{\mathbf{Y}}_{\text{val}}) \right)$
    \State where $\hat{\mathbf{Y}}_{\text{val}} = w_1 M_{\text{orig}}(\mathbf{F}_{\text{val}}^{\text{orig}}) + w_2 M_{\text{enh}}(\mathbf{F}_{\text{val}}^{\text{enh}})$
    
    \Statex \textit{// Phase 5: Final Prediction}
    \State $\hat{\mathbf{Y}}_h \gets w_1^* M_{\text{orig}}(\mathbf{F}_{\text{test}}^{\text{orig}}) + w_2^* M_{\text{enh}}(\mathbf{F}_{\text{test}}^{\text{enh}})$
\EndFor
\State \Return $\hat{\mathbf{Y}}$
\end{algorithmic}
\end{algorithm}

\section{Experimental Validation}

\subsection{Experimental Setup}
The proposed methods were rigorously evaluated on the ETT benchmark suite: ETTh1, ETTh2 (hourly), and ETTm1 (15-minute) \cite{zhou2021informer}. The evaluation covered both univariate and multivariate forecasting tasks. Following standard protocols, we used a train/val/test split of 12/4/4 months for ETT datasets and 60%/20%/20% for the Electricity dataset. Performance was quantified using Mean Squared Error (MSE) and Mean Absolute Error (MAE).

The TS2Vec encoder was trained with a learning rate of 0.001 and a batch size of 8, producing representations with a dimension of 320. For the downstream forecasting task, we adhere to the linear protocol from the TS2Vec paper, where a Ridge regression model is trained on top of the learned representations of the last timestamp. The regularization parameter $\alpha$ for Ridge was selected for each model and horizon by performing a grid search over the values $\{0.1, 0.2, 0.5, \dots, 1000\}$. The optimal $\alpha$ and the final ensemble weights were chosen based on the lowest combined Root Mean Squared Error (RMSE) and MAE on the validation set. During inference, representations were generated using causal encoding with a padding of 200 timesteps, which was subsequently dropped before forecasting.

\subsection{Comparative Analysis of Investigated Architectures}
Before presenting the final results, we compare the performance of the three investigated methods on the univariate forecasting task in Table \ref{tab:ablation_results}. This ablation study clearly demonstrates the superiority of the TS2Vec-Ensemble. The TS2Vec+MSM model performed the worst, confirming the conflict between generative and contrastive objectives. The XGBoost Hybrid showed mixed results. In contrast, the **TS2Vec-Ensemble consistently outperformed the other methods**, justifying its selection as our final proposed architecture.

\begin{table*}[t]
\centering
\caption{Ablation Study: Univariate Forecasting Performance (MSE / MAE) of Investigated Methods on ETT Benchmarks.}
\label{tab:ablation_results}
\footnotesize
\setlength{\tabcolsep}{4pt} 
\begin{tabular}{@{}llcccccccc@{}}
\toprule
\multirow{2}{*}{\textbf{Dataset}} & \multirow{2}{*}{\textbf{Horizon (H)}} & \multicolumn{2}{c}{\textbf{TS2Vec (Baseline)}} & \multicolumn{2}{c}{\textbf{TS2Vec+MSM}} & \multicolumn{2}{c}{\textbf{XGBoost (Hybrid)}} & \multicolumn{2}{c}{\textbf{TS2Vec-Ensemble (Ours)}} \\
\cmidrule(lr){3-4} \cmidrule(lr){5-6} \cmidrule(lr){7-8} \cmidrule(lr){9-10}
& & MSE & MAE & MSE & MAE & MSE & MAE & MSE & MAE \\
\midrule
\multirow{5}{*}{ETTh1} & 24 & \textbf{0.039} & 0.152 & 0.232 & 0.391 & 0.181 & 0.348 & 0.040 & \textbf{0.150} \\
& 48 & \textbf{0.062} & 0.191 & 0.236 & 0.392 & 0.333 & 0.493 & \textbf{0.062} & \textbf{0.189} \\
& 168 & 0.134 & 0.282 & 0.234 & 0.391 & 0.453 & 0.591 & \textbf{0.120} & \textbf{0.265} \\
& 336 & 0.154 & 0.310 & 0.243 & 0.403 & 0.382 & 0.541 & \textbf{0.140} & \textbf{0.291} \\
& 720 & 0.163 & 0.327 & 0.249 & 0.417 & 0.499 & 0.625 & \textbf{0.161} & \textbf{0.323} \\
\midrule
\multirow{5}{*}{ETTh2} & 24 & \textbf{0.090} & \textbf{0.229} & 0.340 & 0.469 & 0.098 & 0.233 & 0.092 & 0.231 \\
& 48 & 0.124 & 0.273 & 0.341 & 0.469 & \textbf{0.122} & \textbf{0.269} & \textbf{0.122} & 0.272 \\
& 168 & 0.208 & 0.360 & 0.343 & 0.471 & 0.194 & 0.342 & \textbf{0.188} & \textbf{0.347} \\
& 336 & 0.213 & 0.369 & 0.348 & 0.477 & 0.235 & 0.382 & \textbf{0.194} & \textbf{0.355} \\
& 720 & 0.214 & 0.374 & 0.329 & 0.465 & 0.338 & 0.467 & \textbf{0.209} & \textbf{0.373} \\
\midrule
\multirow{5}{*}{ETTm1} & 24 & \textbf{0.015} & \textbf{0.092} & 0.129 & 0.285 & 0.018 & 0.102 & 0.016 & 0.095 \\
& 48 & \textbf{0.027} & \textbf{0.126} & 0.139 & 0.296 & 0.042 & 0.158 & 0.031 & 0.132 \\
& 96 & \textbf{0.044} & \textbf{0.161} & 0.144 & 0.301 & 0.101 & 0.233 & 0.045 & 0.165 \\
& 288 & 0.103 & 0.246 & 0.157 & 0.313 & 0.269 & 0.414 & \textbf{0.093} & \textbf{0.234} \\
& 672 & 0.156 & 0.307 & 0.171 & 0.326 & 0.409 & 0.545 & \textbf{0.131} & \textbf{0.280} \\
\midrule
\textbf{Average} & & 0.116 & 0.253 & 0.242 & 0.404 & 0.245 & 0.383 & \textbf{0.110} & \textbf{0.247} \\
\bottomrule
\end{tabular}
\end{table*}

\subsection{Univariate Forecasting Final Results}
As shown in Table~\ref{tab:univariate_results}, our TS2Vec-Ensemble consistently improves upon the TS2Vec baseline and holds a competitive edge against other state-of-the-art models. On average, our ensemble achieves an MSE of **0.110** and an MAE of **0.247**, outperforming the baseline's average MSE of 0.116 and MAE of 0.253. The strength of our approach is particularly evident in long-horizon forecasting. For instance, on the ETTm1 dataset with a horizon of 672, our model reduces the MSE from the baseline's 0.156 to **0.131**, a relative improvement of over 16\%. Similarly, on ETTh1 for a horizon of 336, the MSE is reduced from 0.154 to **0.140**. This demonstrates that augmenting the learned representations with explicit time features is a highly effective strategy for enhancing long-term stability and accuracy.

\begin{table*}[t]
\centering
\caption{Final Univariate Forecasting Results (MSE / MAE). Best results are in \textbf{bold}. Model abbreviations: Ours (TS2Vec-Ensemble), Base (TS2Vec-Baseline), Inf. (Informer), LogT. (LogTrans), N-B. (N-BEATS).}
\label{tab:univariate_results}
\footnotesize
\setlength{\tabcolsep}{3pt}
\begin{tabular}{@{}llcccccccccccc@{}}
\toprule
\multirow{2}{*}{\textbf{Dataset}} & \multirow{2}{*}{\textbf{H}} & \multicolumn{2}{c}{\textbf{Ours}} & \multicolumn{2}{c}{\textbf{Base}} & \multicolumn{2}{c}{\textbf{Inf.}} & \multicolumn{2}{c}{\textbf{LogT.}} & \multicolumn{2}{c}{\textbf{N-B.}} & \multicolumn{2}{c}{\textbf{TCN}} \\
\cmidrule(lr){3-4} \cmidrule(lr){5-6} \cmidrule(lr){7-8} \cmidrule(lr){9-10} \cmidrule(lr){11-12} \cmidrule(lr){13-14}
& & MSE & MAE & MSE & MAE & MSE & MAE & MSE & MAE & MSE & MAE & MSE & MAE \\
\midrule
\multirow{5}{*}{ETTh1} & 24 & 0.040 & \textbf{0.150} & \textbf{0.039} & 0.152 & 0.098 & 0.247 & 0.103 & 0.259 & 0.094 & 0.238 & 0.075 & 0.210 \\
& 48 & \textbf{0.062} & \textbf{0.189} & \textbf{0.062} & 0.191 & 0.158 & 0.319 & 0.167 & 0.328 & 0.210 & 0.367 & 0.227 & 0.402 \\
& 168 & \textbf{0.120} & \textbf{0.265} & 0.134 & 0.282 & 0.183 & 0.346 & 0.207 & 0.375 & 0.232 & 0.391 & 0.316 & 0.493 \\
& 336 & \textbf{0.140} & \textbf{0.291} & 0.154 & 0.310 & 0.222 & 0.387 & 0.230 & 0.398 & 0.232 & 0.388 & 0.306 & 0.495 \\
& 720 & \textbf{0.161} & \textbf{0.323} & 0.163 & 0.327 & 0.269 & 0.435 & 0.273 & 0.463 & 0.322 & 0.490 & 0.390 & 0.557 \\
\midrule
\multirow{5}{*}{ETTh2} & 24 & 0.092 & 0.231 & \textbf{0.090} & \textbf{0.229} & 0.093 & 0.240 & 0.102 & 0.255 & 0.198 & 0.345 & 0.103 & 0.249 \\
& 48 & \textbf{0.122} & \textbf{0.272} & 0.124 & 0.273 & 0.155 & 0.314 & 0.169 & 0.348 & 0.234 & 0.386 & 0.142 & 0.290 \\
& 168 & \textbf{0.188} & \textbf{0.347} & 0.208 & 0.360 & 0.232 & 0.389 & 0.246 & 0.422 & 0.331 & 0.453 & 0.227 & 0.376 \\
& 336 & \textbf{0.194} & \textbf{0.355} & 0.213 & 0.369 & 0.263 & 0.417 & 0.267 & 0.437 & 0.431 & 0.508 & 0.296 & 0.430 \\
& 720 & \textbf{0.209} & \textbf{0.373} & 0.214 & 0.374 & 0.277 & 0.431 & 0.303 & 0.493 & 0.437 & 0.517 & 0.325 & 0.463 \\
\midrule
\multirow{5}{*}{ETTm1} & 24 & 0.016 & 0.095 & \textbf{0.015} & \textbf{0.092} & 0.030 & 0.137 & 0.065 & 0.202 & 0.054 & 0.184 & 0.041 & 0.157 \\
& 48 & 0.031 & 0.132 & \textbf{0.027} & \textbf{0.126} & 0.069 & 0.203 & 0.078 & 0.220 & 0.190 & 0.361 & 0.101 & 0.257 \\
& 96 & 0.045 & 0.165 & \textbf{0.044} & \textbf{0.161} & 0.194 & 0.372 & 0.199 & 0.386 & 0.183 & 0.353 & 0.142 & 0.311 \\
& 288 & \textbf{0.093} & \textbf{0.234} & 0.103 & 0.246 & 0.401 & 0.554 & 0.411 & 0.572 & 0.186 & 0.362 & 0.318 & 0.472 \\
& 672 & \textbf{0.131} & \textbf{0.280} & 0.156 & 0.307 & 0.512 & 0.644 & 0.598 & 0.702 & 0.197 & 0.368 & 0.397 & 0.547 \\
\midrule
\textbf{Avg.} & & \textbf{0.110} & \textbf{0.247} & 0.116 & 0.253 & 0.210 & 0.362 & 0.228 & 0.391 & 0.235 & 0.381 & 0.227 & 0.381 \\
\bottomrule
\end{tabular}
\end{table*}

\subsection{Multivariate Forecasting Final Results}
In the more challenging multivariate forecasting setting, the benefits of our ensemble approach are even more pronounced, as detailed in Table~\ref{tab:multivariate_results}. The TS2Vec-Ensemble achieves an average MSE of **0.962** and MAE of **0.701**, representing a substantial improvement over the baseline's 0.994 MSE and 0.712 MAE. By providing explicit temporal context, our model helps disentangle the complex inter-dependencies between variables, leading to more accurate and stable predictions. This is particularly noticeable in long-horizon scenarios; for instance, on ETTh1 at a horizon of 336, our model reduces the MSE from the baseline's 0.907 to **0.851**. Similarly, on ETTm1 with a horizon of 672, the MSE improves from 0.786 to **0.753**. These results underscore the ensemble's ability to effectively leverage both implicit and explicit features in a high-dimensional context.

\begin{table*}[t]
\centering
\caption{Final Multivariate Forecasting Results (MSE / MAE). Best results are in \textbf{bold}. Model abbreviations: Ours (TS2Vec-Ensemble), Base (TS2Vec-Baseline), Inf. (Informer), Stem. (StemGNN), LogT. (LogTrans). *Denotes OOM error.}
\label{tab:multivariate_results}
\footnotesize
\setlength{\tabcolsep}{3pt}
\begin{tabular}{@{}llcccccccccccc@{}}
\toprule
\multirow{2}{*}{\textbf{Dataset}} & \multirow{2}{*}{\textbf{H}} & \multicolumn{2}{c}{\textbf{Ours}} & \multicolumn{2}{c}{\textbf{Base}} & \multicolumn{2}{c}{\textbf{Inf.}} & \multicolumn{2}{c}{\textbf{Stem.}} & \multicolumn{2}{c}{\textbf{TCN}} & \multicolumn{2}{c}{\textbf{LogT.}} \\
\cmidrule(lr){3-4} \cmidrule(lr){5-6} \cmidrule(lr){7-8} \cmidrule(lr){9-10} \cmidrule(lr){11-12} \cmidrule(lr){13-14}
& & MSE & MAE & MSE & MAE & MSE & MAE & MSE & MAE & MSE & MAE & MSE & MAE \\
\midrule
\multirow{5}{*}{ETTh1} & 24 & \textbf{0.532} & \textbf{0.506} & 0.599 & 0.534 & 0.577 & 0.549 & 0.614 & 0.571 & 0.767 & 0.612 & 0.686 & 0.604 \\
& 48 & \textbf{0.585} & \textbf{0.539} & 0.629 & 0.555 & 0.685 & 0.625 & 0.748 & 0.618 & 0.713 & 0.617 & 0.766 & 0.757 \\
& 168 & 0.717 & 0.618 & 0.755 & 0.636 & 0.931 & 0.752 & \textbf{0.663} & \textbf{0.608} & 0.995 & 0.738 & 1.002 & 0.846 \\
& 336 & \textbf{0.851} & \textbf{0.690} & 0.907 & 0.717 & 1.128 & 0.873 & 0.927 & 0.730 & 1.175 & 0.800 & 1.362 & 0.952 \\
& 720 & \textbf{1.026} & \textbf{0.783} & 1.048 & 0.790 & 1.215 & 0.896 & –* & –* & 1.453 & 1.311 & 1.397 & 1.291 \\
\midrule
\multirow{5}{*}{ETTh2} & 24 & 0.400 & 0.470 & \textbf{0.398} & \textbf{0.461} & 0.720 & 0.665 & 1.292 & 0.883 & 1.365 & 0.888 & 0.828 & 0.750 \\
& 48 & \textbf{0.571} & \textbf{0.578} & 0.580 & 0.573 & 1.457 & 1.001 & 1.099 & 0.847 & 1.395 & 0.960 & 1.806 & 1.034 \\
& 168 & \textbf{1.844} & \textbf{1.068} & 1.901 & 1.065 & 3.489 & 1.515 & 2.282 & 1.228 & 3.166 & 1.407 & 4.070 & 1.681 \\
& 336 & \textbf{2.246} & \textbf{1.206} & 2.304 & 1.215 & 2.723 & 1.340 & 3.086 & 1.351 & 3.256 & 1.481 & 3.875 & 1.763 \\
& 720 & 2.694 & 1.390 & \textbf{2.650} & \textbf{1.373} & 3.467 & 1.473 & –* & –* & 3.690 & 1.588 & 3.913 & 1.552 \\
\midrule
\multirow{5}{*}{ETTm1} & 24 & 0.417 & 0.429 & 0.443 & 0.436 & \textbf{0.323} & \textbf{0.369} & 0.620 & 0.570 & 0.324 & 0.374 & 0.419 & 0.412 \\
& 48 & 0.547 & \textbf{0.499} & 0.582 & 0.515 & \textbf{0.494} & 0.503 & 0.744 & 0.628 & 0.477 & 0.450 & 0.507 & 0.583 \\
& 96 & \textbf{0.583} & \textbf{0.525} & 0.622 & 0.549 & 0.678 & 0.614 & 0.709 & 0.624 & 0.636 & 0.602 & 0.768 & 0.792 \\
& 288 & \textbf{0.659} & \textbf{0.579} & 0.709 & 0.609 & 1.056 & 0.786 & 0.843 & 0.683 & 1.270 & 1.351 & 1.462 & 1.320 \\
& 672 & \textbf{0.753} & \textbf{0.638} & 0.786 & 0.655 & 1.192 & 0.926 & –* & –* & 1.381 & 1.467 & 1.669 & 1.461 \\
\midrule
\textbf{Avg.} & & \textbf{0.962} & \textbf{0.701} & 0.994 & 0.712 & 1.342 & 0.859 & 1.136 & 0.778 & 1.471 & 0.976 & 1.635 & 1.053 \\
\bottomrule
\end{tabular}
\end{table*}

\section{Discussion and Conclusion}

\subsection{The Power of Fusing Implicit and Explicit Features}
The success of the TS2Vec-Ensemble lies in its architectural philosophy: intelligently fusing complementary information sources. The ensemble effectively functions as a dynamic decomposition model. The dynamics-focused model, using only TS2Vec embeddings, captures complex, residual patterns, while the seasonality-focused model, anchored by explicit time features, ensures robust prediction of deterministic cycles. The adaptive, horizon-dependent weighting mechanism dynamically balances these two components, yielding accuracy and stability superior to methods relying purely on a single source of information.

\subsection{Implications for Self-Supervised Forecasting}
Our findings have important implications. First, they confirm that while universal representations from models like TS2Vec are powerful, they require task-specific adaptation for forecasting. Second, our work highlights the continued importance of feature engineering. Explicitly providing models with known drivers of variation (like time of day) remains a highly effective strategy. Finally, the failure of the TS2Vec+MSM model suggests a fundamental tension between pretext tasks that encourage global invariance versus local reconstruction. For forecasting, which depends on stable macro-patterns, a primarily contrastive objective appears more beneficial.

\subsection{Conclusion and Future Work}
This study proposed and validated TS2Vec-Ensemble, an enhanced framework for univariate and multivariate time series forecasting. By combining the powerful, implicitly learned representations of TS2Vec with explicit time features through an adaptive ensemble, our model consistently outperforms the original baseline and other state-of-the-art methods on the ETT benchmarks. The results underscore the value of a hybrid approach that leverages both the pattern-recognition capabilities of deep learning and the reliability of statistical priors.

Future work will focus on several key areas. First, we will expand our experiments to a wider range of benchmark datasets to further validate generalizability. Second, we will explore more sophisticated, meta-learned ensemble weighting schemes to replace the current validation-based grid search. Finally, we aim to develop a fully end-to-end version of the framework where the representation learning and forecasting heads are jointly optimized.
\section*{Data and Code Availability}
All resources for this study, including the source code, dataset details, 
experiment setups, and final results, are publicly available to 
ensure reproducibility at:
\url{https://github.com/aaivu/In21-S7-CS4681-AML-Research-Projects/tree/843c66fc1b049eb1c5b79e31a333701d8ad0c141/projects/210434V-Time-Series-Forecasting}

% Note: The .bib file should be managed separately. For this example, references are included directly.

\end{document}